\documentclass[sigconf,authordraft,review=false,timestamp=false]{acmart}
\AtBeginDocument{%
  }

\copyrightyear{2026}
\acmYear{2026}

\acmConference[SIGIR '26] {Proceedings of the 49th International ACM SIGIR Conference on Research and Development in Information Retrieval}{July 20--24, 2026}{Melbourne, VIC, Australia.}
\acmBooktitle{Proceedings of the 49th International ACM SIGIR Conference on Research and Development in Information Retrieval (SIGIR '26), July 20--24, 2026, Melbourne, VIC, Australia}
\acmISBN{979-8-4007-2599-9/2026/07}
\acmDOI{10.1145/3805712.3809604}
\usepackage{enumitem}
\usepackage{subcaption}
\usepackage{multirow}
\usepackage{color}

\begin{document}

\title{RES-MR: Risk-Aware Reasoning for Explainable and Safe Medication Recommendation}

\author{Cong Wang}
\authornote{Equally contributed.}
\orcid{0009-0009-2993-0248}
\affiliation{%
  \institution{Key Laboratory of Computing Power Network and Information Security, Ministry of Education, Shandong Computer Science Center (National Supercomputer Center in Jinan), Qilu University of Technology (Shandong Academy of Sciences); Shandong Provincial Key Laboratory of Computing Power Internet and Service Computing, Shandong Fundamental Research Center for Computer Science}
  \city{Jinan}
  \state{Shandong}
  \country{China}
}
\email{cong.wang2024@gmail.com}

\author{Jin Li}
\orcid{0000-0001-5737-3594}
\authornotemark[1]
\affiliation{%
  \institution{University of Technology Sydney}
  \city{Sydney}
  \state{NSW}
  \country{Australia}}
\email{jin.li-4@student.uts.edu.au}

\author{Shoujin Wang}
\orcid{0000-0003-1133-9379}
\affiliation{%
  \institution{University of Technology Sydney}
  \city{Sydney}
  \state{NSW}
  \country{Australia}}
\email{shoujin.wang@uts.edu.au}

\author{Yishuo Li}
\orcid{0009-0002-0093-9854}
\affiliation{%
  \institution{Key Laboratory of Computing Power Network and Information Security, Ministry of Education, Shandong Computer Science Center (National Supercomputer Center in Jinan), Qilu University of Technology (Shandong Academy of Sciences); Shandong Provincial Key Laboratory of Computing Power Internet and Service Computing, Shandong Fundamental Research Center for Computer Science}
  \city{Jinan}
  \state{Shandong}
  \country{China}
}
\email{yishuo.li@foxmail.com}

\author{Huilin Gu}
\orcid{0009-0002-6116-2524}
\affiliation{%
  \institution{University of Technology Sydney}
  \city{Sydney}
  \state{NSW}
  \country{Australia}}
\email{huilin.gu@student.uts.edu.au}

\author{Wenpeng Lu}
\orcid{0000-0002-1840-3540}
\authornote{Corresponding author.}
\affiliation{%
  \institution{Key Laboratory of Computing Power Network and Information Security, Ministry of Education, Shandong Computer Science Center (National Supercomputer Center in Jinan), Qilu University of Technology (Shandong Academy of Sciences); Shandong Provincial Key Laboratory of Computing Power Internet and Service Computing, Shandong Fundamental Research Center for Computer Science}
  \city{Jinan}
  \state{Shandong}
  \country{China}
}
\email{wenpeng.lu@qlu.edu.cn}

\renewcommand{\shortauthors}{Cong Wang et al.}

\begin{abstract}
Medication recommendations seek to deliver personalized drug combinations based on patients' clinical information, where ensuring transparent decision-making and personalized safety are of paramount importance. However, existing approaches predominantly operate as ``black boxes'', offering limited interpretability and hindering clinician trust. Furthermore, they typically rely on fixed, patient-agnostic safety constraints, neglecting the heterogeneity of risk tolerance across individuals, which poses significant risks to vulnerable populations. To address these gaps, we propose \textbf{RES-MR}, a novel Risk-aware Reasoning framework for Explainable and Safe Medication Recommendation using LLMs. RES-MR follows a two-stage training paradigm: (1) an explainable clinical reasoning distillation stage to elicit diagnostic capabilities by fine-tuning on synthesized reasoning trajectories, grounded in patient-specific knowledge from graph retrieval; and (2) a risk-aware policy optimization stage to dynamically balance therapeutic efficacy with personalized safety. This stage employs disentangled drug factors to construct patient-specific reference prototypes that capture desired therapeutic drug features and risk features to avoid. These prototypes enable risk boundary calibration and guide the model toward safe recommendations via reward shaping and clipping. Extensive experiments on benchmark datasets demonstrate that RES-MR significantly outperforms state-of-the-art baselines in terms of accuracy, safety, and explainability. Our code and data are available at: https://github.com/wangcong2001/RES-MR.
\end{abstract}

\ccsdesc[500]{Information systems~Recommender systems}
\ccsdesc[500]{Social and professional topics~Medical records}

\keywords{Medication recommendation, Electronic health record, Large language models, Explainability, Medication safety}

\maketitle

\section{Introduction}

Medication recommendation is a pivotal task in clinical decision support systems, aiming to assist physicians in tailoring effective and safe drug combinations for individual patients~\cite{survey, survey2, wang2024hierarchical, wang2020modelling, wang2022exploiting, huang2023dual,WSD_Fanqing,Rumor_zhangqi}. Reliable medication recommender systems hold immense potential to not only mitigate preventable medication errors and reduce adverse patient outcomes but also significantly alleviate the cognitive load on clinicians~\cite{merror3, LEAP, smrec}. However, in real-world clinical practice, this task is substantially more complex than following general recommendations, as accurate and safe medication decisions require jointly considering therapeutic benefit, potential drug-drug interactions, contraindications, and patient-specific context~\cite{ddisurvey}. 
To address this, there is a growing paradigm shift towards utilizing data-driven approaches based on large-scale Electronic Health Records (EHRs)~\cite{ehr3,ehr4, PGbench} that capture the rich longitudinal patient data necessary to model these intricate dependencies, aiming to automate this complex decision-making process.

\begin{figure}
    \centering
    \includegraphics[width=0.85\linewidth]{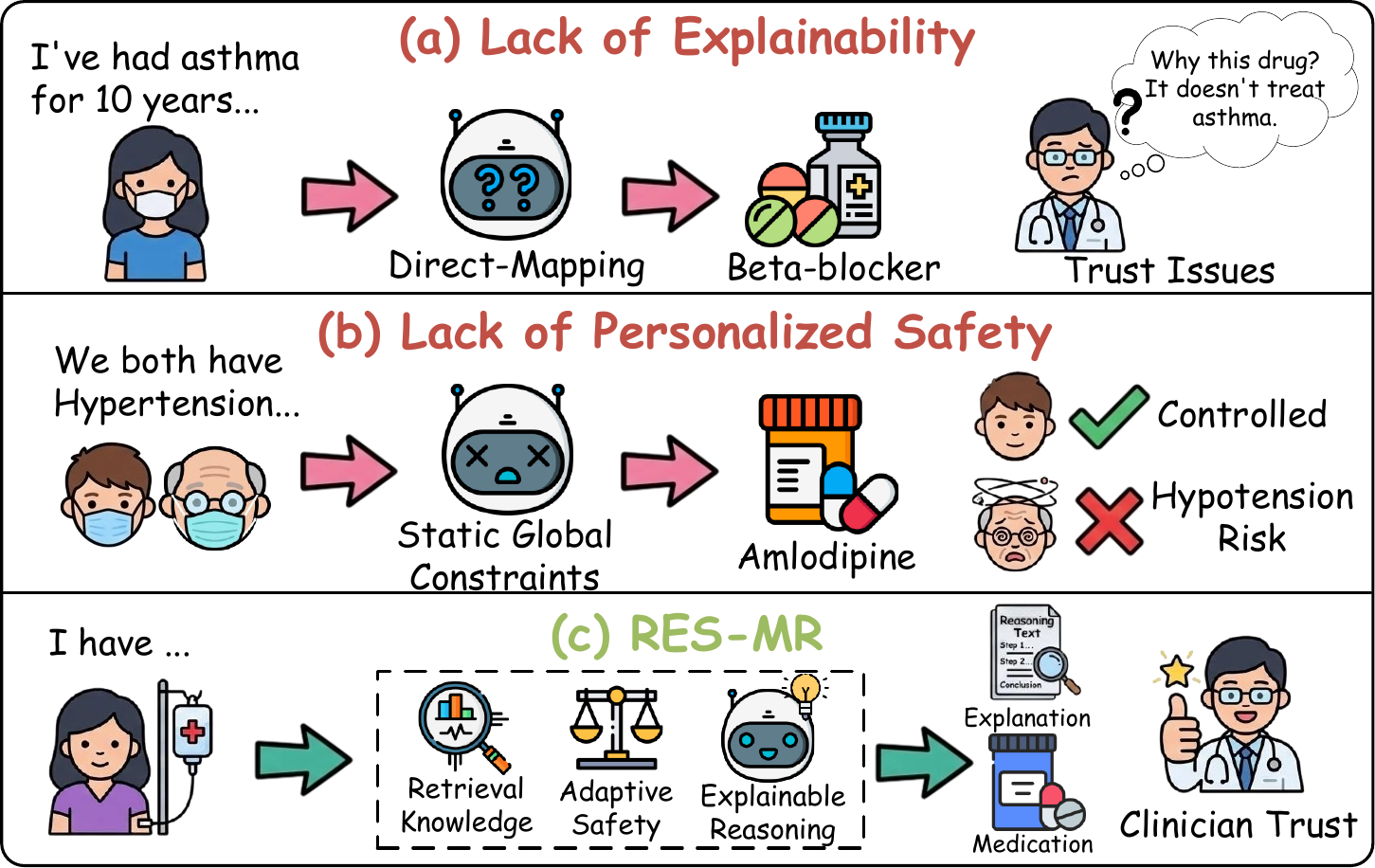}
    \captionsetup{font=small}
    \caption{Examples illustrating two limitations of existing methods. (a) Most RNN/GNN/LLM methods follow a ``direct-mapping'' paradigm and usually output drugs (e.g., a beta-blocker for asthma)  without providing an explicit rationale, leading to trust issues.  (b) Safety-oriented methods use static global constraints, imposing a one-size-fits-all risk tolerance and thus risking vulnerable patients or over-restricting robust ones. (c) RES-MR leverages retrieval-augmented reasoning and personalized boundaries to generate verifiable explanations and safe medication recommendations.}
    \label{fig:example}
\end{figure}

A substantial body of work has investigated deep learning approaches for EHR-based medication recommendation, aiming to infer personalized drug combinations from patients’ structured clinical codes and longitudinal histories.
While early \textit {instance-based methods}~\cite{LEAP,smr2021} focused on extracting features from single visits, the field has largely shifted towards longitudinal modeling.
Specifically, \textit{longitudinal approaches}~\cite{PREMIER, TEMPT} primarily utilize Recurrent Neural Networks (RNNs)~\cite{safedrug, gpsrec} to model long-term disease progression and treatment trajectories by aggregating temporal dependencies from patients' historical records. 
Concurrently, \textit{graph-based models} have emerged to reinforce medication safety and structural understanding. 
These methods employ Graph Neural Networks (GNNs)~\cite{GAMENET, refine} to incorporate external knowledge, ranging from drug-drug interaction (DDI) graphs~\cite{safedrug, molerec, dpg} to molecular structural information, thereby explicitly capturing the complex relational associations among diagnoses, procedures, and medications. 
More recently, \textit{Large Language Models} (LLMs)~\cite{KELLM,LEADER,FLAME,zhou2025large} have introduced a new generative paradigm, leveraging their exceptional natural language understanding and extensive medical knowledge for more comprehensive patient modeling.

Despite these advances, existing methods remain limited by opaque decision processes and reliance on fixed, patient-agnostic safety constraints without considering specific patient conditions.
This oversimplified formulation fails to reflect the complexity of real-world clinical practice, as illustrated in Figure~\ref{fig:example}, leaving two critical challenges (CHs) unaddressed: 

\vspace{0.5em}
\noindent \textbf{\textbullet\hspace{0.5em}CH1:} \textit{How to generate a \textbf{transparent, step-by-step reasoning path} that doctors can verify easily?} As illustrated in Figure~\ref{fig:example} (a), the prevalent ``direct mapping'' paradigm, seen in both discriminative GNNs~\cite{ExpDrug} and generative LLM-based methods~\cite{KELLM, LAMO}, bypasses the explicit reasoning path that should connect symptoms, pathology, and treatment rationale. Although some efforts have attempted to mitigate this opacity via attention mechanisms or aspect-based alignment~\cite{ExpDrug}, they primarily provide post-hoc feature attribution, rather than generating explicit diagnostic logic. They highlight which clinical features contributed to the prediction, but fail to articulate how the conclusion was logically deduced. Such a lack of transparency leads to the dilemma of unverifiable errors. For instance, if a model erroneously recommends a beta-blocker to an asthmatic patient, an opaque system cannot clarify whether the error stems from a misinterpretation of the disease status or neglect of pharmacological contraindications. Without a verifiable reasoning path, clinicians often struggle to validate or trust the system's recommendations in high-stakes scenarios.

\vspace{0.5em}
\noindent \textbf{\textbullet\hspace{0.5em}CH2:} \textit{How to balance risks and benefits under an \textbf{adaptive, patient-specific risk boundary} informed by individual clinical conditions?} As shown in Figure~\ref{fig:example} (b), most methods~\cite{cikm2022, smr2021, gpsrec} enforce safety through static, global constraints such as DDI penalties in the loss function~\cite{safedrug} or molecular-structure-informed representations~\cite{medrec, carmen}, without accounting for heterogeneity in patient-specific risk tolerance. In practice, acceptable risk is highly dependent on individual clinical status. Consider hypertension treatment as an example: a physiologically robust young patient may tolerate a more aggressive strategy to maximize long-term control, whereas a frail older patient with multiple comorbidities requires conservative management to avoid adverse events (e.g., hypotension), even at the cost of less stringent blood pressure control. Due to the overlook of such patient-specific tolerance differences, existing models struggle to optimize utility under the patient-specific risk boundary. As a result, their provided recommendations are either overly conservative for robust patients or potentially risky for vulnerable ones.

\begin{figure*}[ht]
    \centering
    \includegraphics[width=0.90\linewidth]{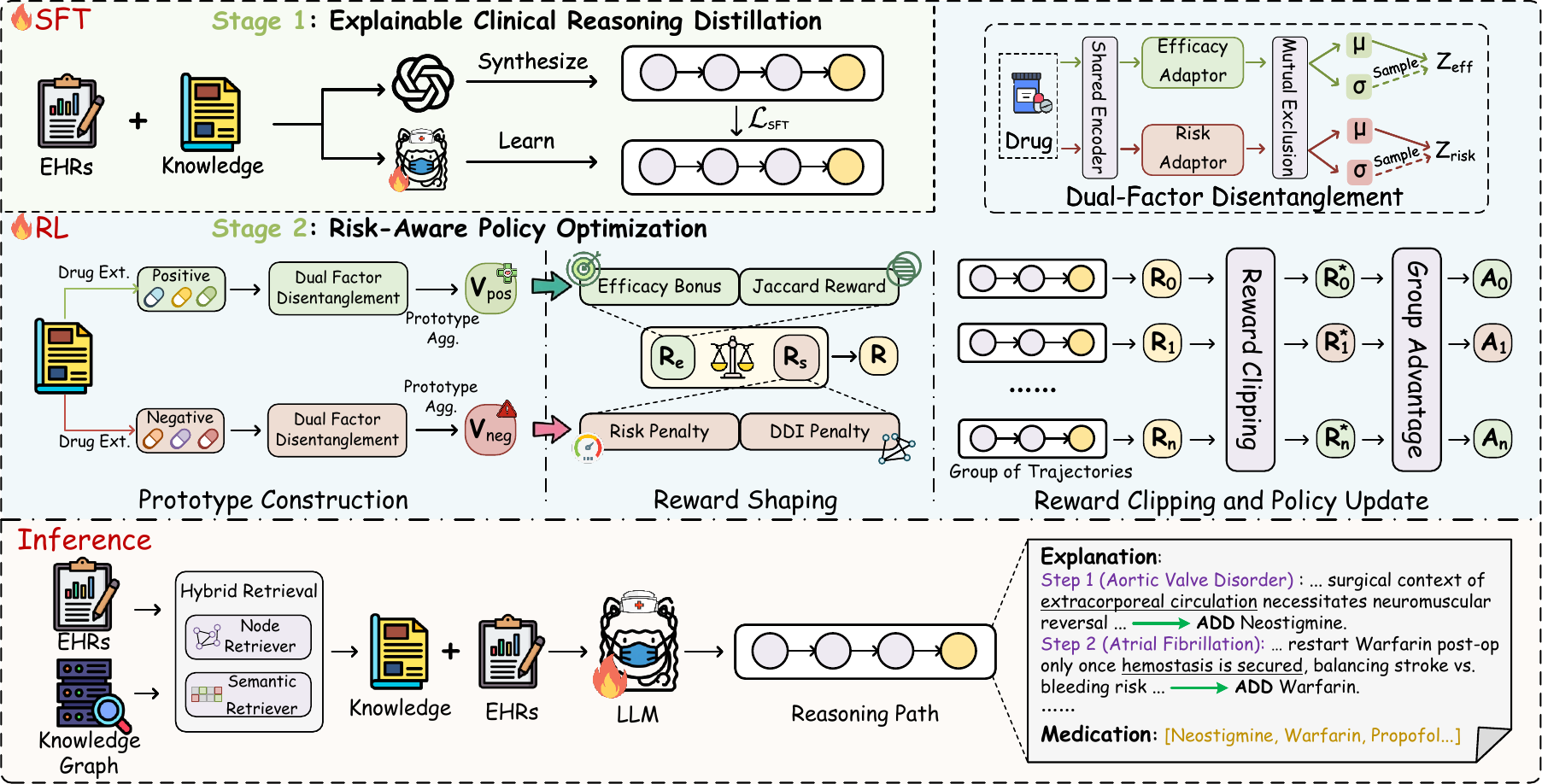}
    \captionsetup{font=small}
    \caption{Overall framework and training/inference pipeline of RES-MR. Stage 1: Given patient EHRs, hybrid retrieval retrieves patient-specific evidence from the knowledge graph; a teacher LLM synthesizes reasoning trajectories to perform Supervise Fine-tuning (SFT) of the LLM, endowing it with strong reasoning capabilities to provide explicit explanations. Stage 2: We further optimize the LLM with Reinforcement Learning (RL). A pretrained dual-factor disentanglement extracts drug efficacy and risk factors to construct patient-specific prototypes, which guide the policy update during RL training via reward shaping and clipping to ensure safety. Inference: The fully trained LLM uses the hybrid retrieval to augment generation with retrieved knowledge and outputs verifiable explanations and safe medication recommendations.}
    \label{fig:frame}
\end{figure*}

To bridge these gaps, we propose \textbf{RES-MR}, a novel Risk-aware \textbf{R}easoning framework for \textbf{E}xplainable and \textbf{S}afe \textbf{M}edication \textbf{R}ecom-mendation using LLMs, as shown in Figure~\ref{fig:frame}.
RES-MR adopts a two-stage training paradigm integrating Supervised Fine-Tuning (SFT)~\cite{lora} and Reinforcement Learning (RL)~\cite{GSPO, EAGRL} to respectively address the aforementioned two challenges.
\textbf{Stage 1: Explainable Clinical Reasoning Distillation (ECRD).} To address CH1 and provide transparent, verifiable decision paths, we apply SFT to endow the LLM with generative reasoning capabilities. First, to ensure the reasoning is grounded in clinical facts rather than hallucinations, we introduce a knowledge graph retrieval module to extract patient-specific evidence. Then, we employ a reasoning distillation strategy, fine-tuning the LLM on high-quality reasoning trajectories to explicitly articulate logical chains as explanations (an example is shown in Figure~\ref{fig:frame}). By connecting symptoms to pharmacological mechanisms based on retrieved evidence, this stage effectively ensures the decision process is explainable and clinically verifiable.
\textbf{Stage 2: Risk-Aware Policy Optimization (RAPO).} To address CH2 and achieve personalized risk-benefit trade-offs, we align the model with patient-specific safety constraints via RL. The core idea is to guide the model using adaptive reward shaping to explore effective treatments without exceeding patient's risk tolerance. To quantify this tolerance, we first design a dual-factor disentanglement module to decompose drug representations into efficacy and risk factors. These factors are then aggregated into patient prototypes, representing desired medication and avoided medication. Leveraging the patient's historical records and constructed prototypes, we can calibrate a personalized risk boundary. Furthermore, we present a reward clipping strategy to prevent potentially imbalanced rewards.
Based on this two-stage training process, RES-MR generates explicit reasoning trajectories as explanations while providing personalized safe medication recommendations.

In summary, the main contributions of this paper are as follows:
\begin{itemize}[leftmargin=9pt]
\item We propose RES-MR, a novel framework for explainable and safe medication recommendation using LLMs. It integrates reasoning generation and safety alignment into a two-stage paradigm, effectively addressing the challenges of opaque decision processes and patient-agnostic safety constraints.
\item In Stage 1, we propose explainable clinical reasoning distillation to ensure transparent decision-making. By integrating the knowledge graph retrieval module with a reasoning distillation strategy, we endow the model with the capability of generating explainable reasoning trajectories grounded in clinical evidence. 

\item In Stage 2, we devise risk-aware policy optimization to achieve personalized risk-benefit trade-offs. We introduce a dual-factor disentanglement module and patient prototypes to quantify personalized risk boundaries, which guide an RL policy via adaptive reward shaping and clipping to ensure safe exploration tailored to individual patient tolerance.
\item Extensive experiments on benchmark datasets demonstrate that RES-MR consistently outperforms state-of-the-art baselines across recommendation accuracy, safety, and explainability metrics, validating its potential for reliable clinical decision support.
\end{itemize}

\section{Related Works}
\textbf{Medication Recommendation.} 
Based on the utilization of clinical data, existing studies can be categorized into instance-based and longitudinal methods. Instance-based methods~\cite{smr2021,LEAP}, such as LEAP~\cite{LEAP}, primarily rely on the current visit information to encode diagnoses through multi-instance multi-label learning, often overlooking the long-term progression of diseases. In contrast, longitudinal methods incorporate patient history to model health trajectories~\cite{sspnet,comrlong,sarmr, gbert, DEPOT, vita, hidr}. For instance, COGNet~\cite{COG} introduces a copy-or-predict mechanism to reuse historical medications, while ARMR~\cite{armr} utilizes segmented temporal learning components to capture evolving patient statuses. Regarding transparency, ExpDrug~\cite{ExpDrug} attempts to map patient features into explainable aspect features to improve model interpretability. Recently, Large Language Models (LLMs) have introduced a generative paradigm to this field~\cite{KELLM,LEADER,LAMO,FLAME}. LAMO~\cite{LAMO} integrates structured codes with unstructured clinical text for the first time.

\noindent\textbf{Safe Medication Recommendation.} 
Mitigating adverse drug events and ensuring safe medication recommendations are paramount objectives in clinical applications~\cite{4sdrug, gpsrec}. Early approaches like SafeDrug~\cite{safedrug} employ dual molecular graph encoders to capture structural drug information and minimize DDI losses, while MoleRec~\cite{molerec} enhances safety by modeling the relationships between critical molecular substructures and patient health conditions. ARC~\cite{ARC} improves safety from a reliability perspective by calibrating confidence for medication recommendations at both the individual-drug and set levels. With the advent of LLMs, recent works have attempted to integrate safety constraints into generative frameworks. For example, KELLM~\cite{KELLM} incorporates safety constraints via LoRA fine-tuning with external knowledge, while FLAME~\cite{FLAME} formulates recommendation as a sequential decision process, utilizing step-wise Group Relative Policy Optimization for fine-grained drug list generation. 
Despite these advancements, existing methods lack verifiable reasoning trajectories and overlook patient-specific risk tolerance, hindering the adaptive balance between efficacy and safety across diverse patients.

\noindent\textbf{Reasoning-enhanced LLM-Based Recommendation.}
Recent advancements have spurred interest in leveraging the reasoning capabilities of Large Language Models (LLMs) to move beyond simple rating prediction toward reasoning-enhanced recommendation~\cite{ChatGPTrec, TALLRec, AgentCF, li2026towards}. While in-context learning strategies like Chain-of-Thought (CoT) have been explored, general-purpose LLMs often lack domain-specific optimization without fine-tuning. Consequently, research has shifted toward fine-tuning paradigms to internalize reasoning patterns~\cite{CoTRec, aclreasoningrec, li2025survey, Reason4Rec}. 
EXP3RT~\cite{EXP3RT} proposes a distillation framework that constructs profiles from reviews to jointly generate rationales and ratings in a single step.
Reason4Rec~\cite{Reason4Rec} introduces a deliberative, multi-stage framework supervised by feedback to enable finer control over reasoning paths.
However, these general-domain methods prioritize user preference alignment, failing to address the complexity of clinical practice where accurate decisions require jointly considering therapeutic benefits, drug-drug interactions, and contraindications.

\section{Problem Formulation}
\textbf{Electronic Health Records (EHRs).}
Let $\mathcal{D}$, $\mathcal{P}$, and $\mathcal{M}$ denote the sets of diagnoses, procedures, and medications, and $|\mathcal{D}|$, $|\mathcal{P}|$, and $|\mathcal{M}|$ represent their sizes.
A patient's EHR is composed of a temporal sequence of visits $\mathcal{V}=[\mathbf{v}_1, \mathbf{v}_2, \dots, \mathbf{v}_T]$, where the $t$-th visit is a tuple $\mathbf{v}_t=(\mathcal{C}_t, \mathbf{n}_t, \mathbf{m}_t)$.
Here, $\mathcal{C}_t = [\mathbf{d}_t, \mathbf{p}_t, \mathbf{a}_t]$ denotes the structured clinical context with the diagnosis set $\mathbf{d}_t \subseteq \mathcal{D}$, the procedure set $\mathbf{p}_t \subseteq \mathcal{P}$, and patient's demographic attributes $\mathbf{a}_t$.
In contrast, $\mathbf{n}_t$ is the unstructured clinical note, and $\mathbf{m}_t \subseteq \mathcal{M}$ denotes the medication set prescribed in the visit. 
In our framework, the structured context $\mathcal{C}_t$, the clinical note $\mathbf{n}_t$, and the medications in $\mathbf{m}_t$ are all described in natural language.

\noindent\textbf{Medical Knowledge Graph.}
To support risk-aware reasoning, we utilize an external medical knowledge graph $\mathcal{G}_{kg}=(\mathcal{E},\mathcal{R},\mathcal{T}_{kg})$.
Here, $\mathcal{E}$ represents the set of medical entities, and $\mathcal{R}$ denotes various clinical relations (e.g., \textit{Indication} and \textit{Contraindication}).
$\mathcal{T}_{kg}$ consists of triplets $(h,r,t)$ describing the connections between entities, where $h,t\in\mathcal{E}$ denote the head and tail entities, and $r\in\mathcal{R}$ denotes the relation between them.

\noindent\textbf{Task Definition.}
Consider a patient with a sequence of visits $\mathcal{V}$.
For the $t$-th visit, given the historical visits $\mathbf{v}_{1:t-1}=[\mathbf{v}_1, \dots, \mathbf{v}_{t-1}]$, the current visit context $(\mathcal{C}_t, \mathbf{n}_t)$, and the external knowledge $\mathcal{G}_{kg}$, our goal is to train an LLM $\pi_\theta$, such that it can explicitly generate:
(1) a reasoning trajectory $\mathcal{T}^{\text{reason}}_t$ that explains the clinical decision process; and
(2) a safe medication set $\hat{\mathbf{m}}_t$ for the current visit.

\section{Methodology}

As illustrated in Figure~\ref{fig:frame}, RES-MR aims to generate explainable and safe medication recommendations by leveraging the exceptional reasoning capabilities of LLMs. It adopts a two-stage training paradigm. The first stage, \textbf{Explainable Clinical Reasoning Distillation (ECRD)}, focuses on addressing the opacity of decision-making. By integrating knowledge graph retrieval with a reasoning distillation strategy, it teaches the model via SFT to articulate step-by-step reasoning trajectories grounded in clinical evidence. Building upon this foundation, the second stage, \textbf{Risk-Aware Policy Optimization (RAPO)}, is devised to enforce patient-specific risk-benefit trade-offs through a reward-guided RL process. To quantify personalized risk boundaries, it first decomposes drug representations into efficacy and risk factors and aggregates them into prototypes that capture desired and avoided medications. Using these prototypes together with the patient's historical records, we calibrate a personalized risk boundary. Finally, we optimize the policy via adaptive reward shaping and clipping, guiding the model to explore effective treatments while strictly suppressing boundary violations.

\subsection{Explainable Clinical Reasoning Distillation}
To enable the generation of verifiable reasoning paths for the LLM $\pi_{\theta}$, this stage comprises two steps: (1) knowledge graph retrieval to construct evidence that grounds the reasoning process in objective medical facts; and (2) supervised fine-tuning on synthesized reasoning trajectories to guide the model to explicitly articulate diagnostic logic as explanations.

\subsubsection{Knowledge Graph Retrieval}
For a given patient at the current visit, we perform both \textit{node} and \textit{semantic} retrieval on the Medical Knowledge Graph to construct the patient-specific knowledge context~\cite{graphrag1, graphrag2, graphrag3}. 
First, to capture explicit structural relations, we employ \textit{node retrieval} by querying the KG with the current structured context $\mathcal{C}_t$ (e.g., diagnoses and procedures) and collecting one-hop neighboring triples of these entities to form $\mathcal{G}_{\text{node}}$. 
Complementarily, to uncover implicit semantic associations, we perform \textit{semantic retrieval}. We embed the current note $\mathbf{n}_t$ into a dense vector space and retrieve the most relevant KG contexts via cosine similarity search, yielding the semantic knowledge context $\mathcal{G}_{\text{sem}}$. 
Finally, we merge the two streams into a unified knowledge context $\mathcal{G}_{\text{ret}}=\mathcal{G}_{\text{node}}\cup\mathcal{G}_{\text{sem}}$, which not only provides patient-specific evidence for reasoning but also serves as the knowledge foundation for patient prototype construction.

\subsubsection{Supervised Fine-tuning}
In practice, clinicians formulate prescriptions following a strong rationale from the patient’s specific diagnoses. To endow the model with such diagnostic reasoning capabilities, we leverage GPT-5 to synthesize high-quality reasoning trajectories $\mathcal{T}^{\text{reason}}_t$~\cite{deepseekr1, distillation1} for model fine-tuning. 
Specifically, GPT-5 is prompted with a comprehensive patient context, including the structured context $\mathcal{C}_t$ and note $\mathbf{n}_t$, the retrieved knowledge $\mathcal{G}_{\text{ret}}$ for grounding, and the ground-truth medication $\mathbf{m}_t$. 
The resulting trajectory provides step-by-step, diagnosis-aware explanations that connect patient evidence and symptoms to pharmacological mechanisms and medication actions. 
A clear example is shown in Figure~\ref{fig:frame}.
We fine-tune the LLM $\pi_\theta$ using the synthesized dataset $\mathcal{D}$ with low-rank adaptation~\cite{lora}. Each instance includes the patient history $\mathbf{v}_{1:t-1}$, current visit inputs $(\mathcal{C}_t,\mathbf{n}_t)$, and retrieved knowledge $\mathcal{G}_{\text{ret}}$. 
We minimize the negative log-likelihood of generating $y$:
\begin{equation}
    \mathcal{L}_{\text{SFT}} = 
    - \mathbb{E}_{(x,y)\sim \mathcal{D}}
    \left[
    \sum_{j=1}^{|y|} \log \pi_\theta\!\left(
    y_j \mid x, y_{<j}
    \right)\right],
\end{equation}
where we define the input $x = (\mathbf{v}_{1:t-1}, \mathcal{C}_t, \mathbf{n}_t, \mathcal{G}_{\text{ret}})$ and $y$ represents the concatenated token sequence of the reasoning path $\mathcal{T}^{\text{reason}}_t$ and the final result $\mathbf{m}_t$. 
By learning this structured generation pattern, the LLM acquires diagnosis-aware reasoning behaviors and a strong, format-stable initialization for the RAPO stage.

\subsection{Risk-Aware Policy Optimization}
While SFT establishes fundamental clinical reasoning, its supervision objective overlooks the efficacy-safety trade-off. Consequently, the model struggles to satisfy personalized risk constraints while maintaining accuracy. Therefore, we propose Risk-Aware Policy Optimization (RAPO), which incorporates the trade-off via specially devised reinforcement learning~\cite{rl1, dapo}. The core objective is to guide the model to explore more effective medication combinations while adhering to personalized risk boundaries. To achieve this, we first pretrain a dual-factor disentanglement module to decompose drug representations into efficacy and risk factors, making both signals explicitly quantifiable for reward computation. Based on retrieved knowledge, we then construct patient prototypes that summarize desired and avoided medications. Using these prototypes together with the patient’s medication history, we calibrate a patient-specific risk boundary. In this stage, these signals are used for reward shaping to balance efficacy and safety. We further apply reward clipping to prevent potentially imbalanced rewards.

\subsubsection{Dual-Factor Disentanglement}
To make efficacy and risk signals quantifiable for reward measurement, we introduce a \textit{Dual-Factor Disentanglement} module and pretrain it before RL~\cite{lu2020concept}. 
For each drug $m$, we obtain its semantic embedding $\mathbf{e}_m$ from an LLM encoder. We then encode $\mathbf{e}_m$ with a shared encoder and two factor-specific adaptors, i.e., $\mathbf{h}=f_{\text{share}}(\mathbf{e}_m)$, $\mathbf{h}_{\text{eff}}=f_{\text{eff}}(\mathbf{h})$, and $\mathbf{h}_{\text{risk}}=f_{\text{risk}}(\mathbf{h})$.
To suppress shared information across factors, we introduce a \textit{Mutual Exclusion Block}~\cite{exclusion} prior to the variational projection:
\begin{equation}
    \mathbf{h}_{\text{eff}}^{\prime} = \mathbf{h}_{\text{eff}} - \alpha \cdot \sigma(W_{g}[\mathbf{h}_{\text{eff}}; \mathbf{h}_{\text{risk}}]) \odot \text{MLP}(\mathbf{h}_{\text{risk}}),
\end{equation}
where $\sigma$ is the sigmoid function, $W_g$ is a learnable weight matrix, and $\alpha$ is a learnable parameter. A symmetric operation is also applied to $\mathbf{h}_{\text{risk}}^{\prime}$. 
For $k\in\{\text{eff},\text{risk}\}$, we project $\mathbf{h}_k^{\prime}$ to $(\boldsymbol{\mu}_k,\boldsymbol{\sigma}_k)$ and adopt the reparameterization trick~\cite{VAE}: $\mathbf{z}_k = \boldsymbol{\mu}_k + \boldsymbol{\epsilon} \odot \boldsymbol{\sigma}_k$, where $\boldsymbol{\epsilon} \sim \mathcal{N}(\mathbf{0}, \mathbf{I})$.
We align each factor with clinical semantics using KG relations. Specifically, the positive concept set $T_{k}^{+}$ is constructed from $\mathcal{T}_{kg}$: for efficacy ($k=\text{eff}$), positives are clinical concepts linked by \textit{Indication}; for risk ($k=\text{risk}$), positives are those linked by \textit{Contraindication}. Negatives $T_k^{-}$ are randomly sampled from other concepts. 
For each factor $k$, we pull $\mathbf{z}_k$ close to $T_k^+$ and push it away from $T_k^-$: 
\begin{equation}
    \mathcal{L}_{\text{align}}^{(k)} = -\sum_{t \in T_{k}^{+}}\log \sigma(s(\mathbf{z}_{k}, t)) - \sum_{t^{\prime} \in T_{k}^{-}}\log(1 - \sigma(s(\mathbf{z}_{k}, t^{\prime}))),
\end{equation}
where $s(\cdot,\cdot)$ denotes cosine similarity. The overall alignment loss is $\mathcal{L}_{\text{align}} = \mathcal{L}_{\text{align}}^{(\text{eff})} + \mathcal{L}_{\text{align}}^{(\text{risk})}$.
To promote independence and robustness, we impose an orthogonality constraint and the KL regularizer~\cite{vibsuevey}:
\begin{equation}
\mathcal{L}_{\text{orth}} = \left\|\mathrm{Cov}\!\left(\boldsymbol{\mu}_{\text{eff}}, \boldsymbol{\mu}_{\text{risk}}\right)\right\|_{F}^{2},
\end{equation}
\begin{equation}
\mathcal{L}_{\text{KL}} = \sum_{k \in \{\text{eff}, \text{risk}\}}
D_{\text{KL}}\!\left(q_{\phi}(\mathbf{z}_k\mid \mathbf{e}_m) \,\|\, \mathcal{N}(\mathbf{0}, \mathbf{I})\right),
\end{equation}
where $\mathrm{Cov}$ is covariance matrix and $D_{\text{KL}}$ is KL divergence. We parameterize variational posterior: $q_{\phi}(\mathbf{z}_k| \mathbf{e}_m)=\mathcal{N}(\boldsymbol{\mu}_k,\ \mathrm{diag}(\boldsymbol{\sigma}_k^{2}))$. 
The final objective is $\mathcal{L}_{\text{total}} = \mathcal{L}_{\text{align}} + \lambda_{\text{orth}} \mathcal{L}_{\text{orth}} + \beta_{\text{KL}} \mathcal{L}_{\text{KL}}$. After convergence, we use the frozen $\boldsymbol{\mu}_{\text{eff}}$ and $\boldsymbol{\mu}_{\text{risk}}$ as deterministic drug factors for the following optimization.

\subsubsection{Prototype Construction}
Given a patient and the current visit, this module constructs patient-specific prototypes as reference points~\cite{prototype1} for reward computation in RAPO.
Leveraging the pretrained disentangled factors and retrieved knowledge context $\mathcal{G}_{\text{ret}}$, we derive these prototypes to capture the desired therapeutic features and the risk features to avoid. 
Specifically, based on medical relations in the Knowledge Graph (e.g., indications and contraindications), we categorize drug entities from $\mathcal{G}_{\text{ret}}$ into an indication set $\mathcal{M}_{\text{pos}}$ and a contraindication set $\mathcal{M}_{\text{neg}}$. To ensure valid prototype construction, we require both sets to be non-empty for each patient. Subsequently, utilizing the frozen encoder, we aggregate the corresponding disentangled factors of the drugs within each set. The \textit{positive prototype} $\mathbf{v}_{\text{pos}}$ characterizes the desired therapeutic direction by aggregating the efficacy factors of the indicated drugs:
\begin{equation}
    \mathbf{v}_{\text{pos}} = \frac{1}{|\mathcal{M}_{\text{pos}}|}\sum_{m \in \mathcal{M}_{\text{pos}}} \boldsymbol{\mu}_{\text{eff}}(m).
\end{equation}
Conversely, the \textit{negative prototype} $\mathbf{v}_{\text{neg}}$ captures the high-risk profiles that must be strictly avoided, obtained by aggregating the risk factors of the contraindicated drugs:
\begin{equation}
    \mathbf{v}_{\text{neg}} = \frac{1}{|\mathcal{M}_{\text{neg}}|}\sum_{m \in \mathcal{M}_{\text{neg}}} \boldsymbol{\mu}_{\text{risk}}(m).
\end{equation}

\subsubsection{Reward Shaping}
To guide the model toward optimizing the trade-off between therapeutic efficacy and personalized safety, we design a composite reward function that combines efficacy rewards with safety costs under a patient-specific risk boundary. 
Prior to quantifying safety costs, we first calibrate a risk boundary $B$ to define the maximum risk level a patient can tolerate. 
Using historical medications $\mathcal{M}_{\text{hist}} = \bigcup_{i < t} \mathbf{m}_i$ as an empirical proxy~\cite{proxy1, proxy2} of patient-specific risk exposure under prior clinical supervision, we define $B$ as the $\delta$-quantile of risk similarities between historical drugs and the negative prototype $\mathbf{v}_{\text{neg}}$:
\begin{equation}
\label{eq:bt}
    B = \inf \left\{ b \in \mathbb{R} \;\middle|\; \frac{1}{|\mathcal{M}_{\text{hist}}|}\sum_{m \in \mathcal{M}_{\text{hist}}} \mathbb{I}\!\left(s(\boldsymbol{\mu}_{\text{risk}}(m), \mathbf{v}_{\text{neg}}) \le b\right) \ge \delta \right\},
\end{equation}
where $s(\cdot,\cdot)$ denotes cosine similarity and $\delta \in [0, 1]$ controls the strictness. A larger $B$ implies higher tolerance, allowing broader exploration, whereas a smaller $B$ imposes tighter constraints for vulnerable patients. The final reward $R$ is synthesized by subtracting a weighted safety cost from the efficacy reward:
\begin{equation}
    R = (1 - \rho_t) R_{e} - \rho_t R_{s}.
\end{equation}
The \textit{efficacy reward} $R_{e}$ encourages clinically accurate prescriptions by combining the Jaccard similarity with the ground truth $\mathbf{m}^*$ and an efficacy bonus based on the positive prototype $\mathbf{v}_{\text{pos}}$:
\begin{equation}
    R_{e} = \lambda_j \frac{|\hat{\mathbf{m}} \cap \mathbf{m}^*|}{|\hat{\mathbf{m}} \cup \mathbf{m}^*|} + \lambda_e \frac{1}{|\hat{\mathbf{m}}|} \sum_{d \in \hat{\mathbf{m}}} \max\left(0, s(\boldsymbol{\mu}_{\text{eff}}(d), \mathbf{v}_{\text{pos}})\right).
\end{equation}
This efficacy bonus ensures the model receives positive feedback for functionally valid drugs aligned with the desired therapeutic direction, even if they differ from the exact labels. Conversely, the \textit{safety cost} $R_{s}$ penalizes iatrogenic risks, aggregating Drug-Drug Interaction (DDI) rates and violations of the calibrated risk boundary:
\begin{equation}
    R_{s} = \lambda_{d} \text{DDI}(\hat{\mathbf{m}}) + \lambda_{r} \frac{1}{|\hat{\mathbf{m}}|} \sum_{d \in \hat{\mathbf{m}}} 
    \max\left(0,  s(\boldsymbol{\mu}_{\text{risk}}(d), \mathbf{v}_{\text{neg}}) - B \right).
\end{equation}
The second term penalizes only the amount by which a drug's risk similarity to $\mathbf{v}_{\text{neg}}$ exceeds the boundary $B$. 
Crucially, any format violation or hallucinated drug is assigned a fixed penalty ($R=-100$) and treated as an invalid sample, bypassing the computation of $R_e$ and $R_s$. 
Finally, to prevent early convergence to overly conservative behaviors, we introduce an adaptive gating coefficient $\rho_t$ derived from the moving average Jaccard performance $\bar{J}_t$:
\begin{equation}
    \rho_t = \sigma(\bar{J}_t - \tau_t),
\end{equation}
where $\tau_t$ linearly increases during training. This mechanism ensures that strict safety constraints intervene only after the model acquires sufficient clinical reasoning accuracy.

\subsubsection{Reward Clipping and Policy Update}
With the established reward shaping, we may still encounter a critical risk: high efficacy rewards could potentially overshadow safety costs, encouraging the model to trade safety for performance (i.e., reward hacking~\cite{rewardhacking3, rewardhacking2}). 
To address this, we introduce a reward clipping strategy that suppresses ``high-reward but unsafe'' trajectories upon boundary violations, while allowing a higher upper bound for safe trajectories, thereby encouraging safe exploration and conservative updates, respectively~\cite{EAGRL, entropyrl}.
Specifically, we compute the difference between the personalized risk boundary $B$ and the maximum risk similarity among the generated medications, and use it to adaptively clip the raw reward $R$:
\begin{equation}
    \Delta_t = B - \max_{d \in \hat{\mathbf{m}}} s(\boldsymbol{\mu}_{\text{risk}}(d), \mathbf{v}_{\text{neg}}).
\end{equation}
A positive $\Delta_t$ indicates the medication set is within the safe region, while a non-positive $\Delta_t$ signals a boundary violation. 
Based on the sign of $\Delta_t$, we obtain the final reward $R_{\text{total}}$ by clipping the raw shaped reward $R$, where the clipping bounds are computed from the group-wise mean $\mu_R$ and standard deviation $\sigma_R$ over the $G$ sampled trajectories for the same input context:
\begin{equation}
    R_{\text{total}} = \begin{cases} 
    \text{clip}(R, \mu_R - \sigma_R, \mu_R + 2\sigma_R) & \text{if } \Delta_t > 0 \\ 
    \text{clip}(R, \mu_R - 2\sigma_R, \mu_R + \sigma_R) & \text{if } \Delta_t \le 0
    \end{cases}.
\end{equation}
Here $\text{clip}(x, a, b)=\min(\max(x,a),b)$ truncates the reward into $[a,b]$, preventing extreme reward values from dominating policy updates and stabilizing optimization.
When $\Delta_t>0$, we relax the upper bound to allow higher rewards for safe trajectories. When $\Delta_t\le 0$, we tighten the upper bound and relax the lower bound to suppress unsafe high-reward trajectories, mitigating reward hacking and inducing conservative updates.

Finally, to ensure stable convergence under these dynamic constraints, we employ Group Sequence Policy Optimization~\cite{GSPO}. 
Concretely, we sample an input context $x \sim \mathcal{D}$. For each $x$, we draw $G$ candidate sequences $\{y_i\}_{i=1}^{G}\sim \pi_{\theta_{\text{old}}}(\cdot \mid x)$, where each $y_i$ is a complete generation containing the reasoning trajectory and the medication set. We maximize the following objective:
\begin{equation}
\mathcal{J}(\theta) = 
\mathbb{E}
\left[ \frac{1}{G} \sum_{i=1}^G \min \left( s_i(\theta) \widehat{A}_i, \text{clip} \left( s_i(\theta), 1-\varepsilon, 1+\varepsilon \right) \widehat{A}_i \right) \right],
\end{equation}
where $\widehat{A}_i$ is the standardized advantage derived from $R_{\text{total}}$. The importance ratio $s_i(\theta)$ is defined based on sequence likelihood:
\begin{equation}
    s_i(\theta) = \left( \frac{\pi_{\theta}(y_i|x)}{\pi_{\theta_{\text{old}}}(y_i|x)} \right)^{\frac{1}{|y_i|}}.
\end{equation}
This length normalization keeps the importance ratio comparable across varying response lengths, which helps stabilize training.

\subsection{Inference}
During inference, we perform knowledge graph retrieval to obtain $\mathcal{G}_{\text{ret}}$ for the current visit. 
We form the input context as $x=(\mathbf{v}_{1:t-1}, \mathcal{C}_t, \mathbf{n}_t, \mathcal{G}_{\text{ret}})$ and autoregressively generate a serialized output sequence:
\begin{equation}
\hat{y} \sim \pi_\theta(\cdot \mid x).
\end{equation}
The generated $\hat{y}$ for visit $t$ consists of an explicit reasoning trajectory $\hat{\mathcal{T}}^{\text{reason}}_t$ followed by the recommended medication set $\hat{\mathbf{m}}_t$.

\section{Experiments}
\subsection{Experimental Setup}
\textbf{Datasets.} 
We conduct experiments on MIMIC-III~\cite{mimic3}, MIMIC-IV~\cite{mimic4}, and eICU~\cite{eICU}. MIMIC-III and MIMIC-IV are large-scale databases comprising deidentified critical care data from the Beth Israel Deaconess Medical Center. eICU is a multi-center ICU EHR database with deidentified patient records across hospitals. To better reflect real-world clinical practice, we retain the complete set of diagnoses, procedures, and medications without truncation. Additionally, for the MIMIC-III and MIMIC-IV datasets, we only include admissions that have available clinical notes. Regarding text processing, we follow the common practice~\cite{FLAME,LAMO} and extract specific segments from these notes, including history of present illness, past medical history, allergies, and medications on admission. For data standardization, we unify the diagnosis and procedure codes by mapping them to the ICD-9 standard~\cite{ICD9CM}, utilizing their descriptions as model inputs. Similarly, medications are mapped to DrugBank IDs~\cite{drugbank}, with their corresponding generic names to unify diverse formulations. Furthermore, following established protocols~\cite{safedrug, mr-dtr}, we construct the drug-drug interaction matrix $\mathcal{M}_{ddi}$ based on interaction pairs extracted from the DrugBank database~\cite{drugbank}. We split all three datasets into training, validation, and test sets at the patient level with a 4:1:1 ratio. Retaining the full medication set enlarges the label space, making results not directly comparable to prior studies with different preprocessing. 
The detailed statistics of the processed datasets are summarized in Table~\ref{tab:statistics}.

\begin{table}[t]
\renewcommand{\arraystretch}{0.92}
\centering
\caption{Statistics on MIMIC-III, MIMIC-IV, and eICU Datasets}
\label{tab:statistics}
\resizebox{\columnwidth}{!}{
\begin{tabular}{l|c|c|c}
\toprule
\textbf{Items} & \textbf{MIMIC-III} & \textbf{MIMIC-IV} & \textbf{eICU} \\
\midrule
\# of visits / \# of patients & 12,433 / 4,785 & 156,315 / 46,811 & 18,636 / 8,544 \\
dis. / proc. / med. space size & 4,316 / 1,330 / 593 & 19,812 / 10,859 / 743 & 1,562 / 937 / 152\\
avg. / max \# of visits & 5.68 / 29 & 11.07 / 70 & 2.18 / 9 \\
avg. / max \# of dis. per visit & 14.68 / 39 & 13.45 / 56 & 4.79 / 76 \\
avg. / max \# of proc. per visit & 4.69 / 32 & 2.75 / 34 & 8.03 / 118 \\
avg. / max \# of med. per visit & 14.78 / 67 & 14.24 / 128 & 10.35 / 51 \\
\bottomrule
\end{tabular}
}
\vspace{-2mm}
\end{table}

\begin{table*}[t]
\renewcommand{\arraystretch}{0.89}
\centering
\small
\setlength{\tabcolsep}{2pt} 
\caption{Comparison of accuracy, safety, and explainability performance with baselines on the MIMIC-III, MIMIC-IV, and eICU datasets. $\uparrow$ and $\downarrow$ indicate higher and lower is better, respectively. The best and second-best results are highlighted in \textbf{bold} and \underline{underlined}. * denotes statistical significance ($p < 0.05$) by a paired $t$-test.}
\vspace{-2mm}
\label{tab:overall}
\begin{tabular}{l|l|ccccccc}
\hline
Dataset & Methods
& Jaccard $\uparrow$
& F1 $\uparrow$
& DDI $\downarrow$
& RIS $\downarrow$
& BLEURT$^{1}$ $\uparrow$
& $F_{AS}$ $\uparrow$
& $F_{AE}$ $\uparrow$ \\
\hline

\multirow{15}{*}{MIMIC-III}
& LEAP & 0.1934 $\pm$ 0.0023 & 0.3285 $\pm$ 0.0032 & 0.1341 $\pm$ 0.0012 & 0.1433 $\pm$ 0.0023 & - & 0.4005 $\pm$ 0.0029 & - \\
& COGNet & 0.2381 $\pm$ 0.0019 & 0.3628 $\pm$ 0.0032 & 0.1466 $\pm$ 0.0013 & 0.1473 $\pm$ 0.0021 & - & 0.4444 $\pm$ 0.0015 & - \\
& RAREMed & 0.2244 $\pm$ 0.0046 & 0.3414 $\pm$ 0.0012 & 0.1482 $\pm$ 0.0046 & 0.1547 $\pm$ 0.0038 & - & 0.4243 $\pm$ 0.0026 & - \\
& ARMR & 0.2356 $\pm$ 0.0055 & 0.3599 $\pm$ 0.0035 & 0.1536 $\pm$ 0.0025 & 0.1613 $\pm$ 0.0035 & - & 0.4400 $\pm$ 0.0040 & - \\
& LAMO & 0.2512 $\pm$ 0.0003 & 0.3707 $\pm$ 0.0027 & 0.1124 $\pm$ 0.0023 & 0.1427 $\pm$ 0.0027 & - & 0.4585 $\pm$ 0.0015 & - \\
\cline{2-9}
& SafeDrug & 0.2210 $\pm$ 0.0023 & 0.3469 $\pm$ 0.0012 & 0.1054 $\pm$ 0.0014 & 0.1484 $\pm$ 0.0015 & - & 0.4285 $\pm$ 0.0012 & - \\
& MoleRec & 0.2454 $\pm$ 0.0052 & 0.3770 $\pm$ 0.0038 & 0.1044 $\pm$ 0.0007 & 0.1401 $\pm$ 0.0005 & - & 0.4595 $\pm$ 0.0034 & - \\
& KELLM & 0.2291 $\pm$ 0.0031 & 0.3507 $\pm$ 0.0029 & 0.1318 $\pm$ 0.0025 & 0.1403 $\pm$ 0.0040 & - & 0.4341 $\pm$ 0.0019 & - \\
& ARC & 0.2377 $\pm$ 0.0012 & 0.3610 $\pm$ 0.0038 & 0.1167 $\pm$ 0.0033 & 0.1454 $\pm$ 0.0014 & - & 0.4453 $\pm$ 0.0024 & - \\
& FLAME & \underline{0.2601 $\pm$ 0.0031} & \underline{0.3880 $\pm$ 0.0033} & 0.1291 $\pm$ 0.0009 & 0.1374 $\pm$ 0.0013 & - & \underline{0.4717 $\pm$ 0.0026} & - \\
\cline{2-9}
& ExpDrug & 0.2326 $\pm$ 0.0014 & 0.3610 $\pm$ 0.0022 & 0.1047 $\pm$ 0.0017 & 0.1397 $\pm$ 0.0028 & - & 0.4436 $\pm$ 0.0016 & - \\
& Reason4Rec & 0.1529 $\pm$ 0.0036 & 0.2519 $\pm$ 0.0022 & 0.1512 $\pm$ 0.0064 & \underline{0.1311 $\pm$ 0.0026} & 0.4052 $\pm$ 0.0011 & 0.3276 $\pm$ 0.0022 & 0.2700 $\pm$ 0.0015 \\
& EXP3RT & 0.2150 $\pm$ 0.0028 & 0.3402 $\pm$ 0.0026 & 0.1589 $\pm$ 0.0014 & 0.1531 $\pm$ 0.0035 & 0.3752 $\pm$ 0.0005 & 0.4178 $\pm$ 0.0026 & 0.3191 $\pm$ 0.0015 \\
\cline{2-9}
& RES-MR-3B & 0.2407 $\pm$ 0.0044 & 0.3681 $\pm$ 0.0017 & \underline{0.1041 $\pm$ 0.0019} & 0.1342 $\pm$ 0.0006 & \underline{0.4390 $\pm$ 0.0030} & 0.4524 $\pm$ 0.0024 & \underline{0.3595 $\pm$ 0.0017} \\
& RES-MR-8B & \textbf{0.2803 $\pm$ 0.0046}* & \textbf{0.4179 $\pm$ 0.0016}* & \textbf{0.1031 $\pm$ 0.0006}* & \textbf{0.1267 $\pm$ 0.0045}* & \textbf{0.4538 $\pm$ 0.0026}* & \textbf{0.5007 $\pm$ 0.0020}* & \textbf{0.3946 $\pm$ 0.0017}* \\
\hline

\multirow{15}{*}{MIMIC-IV}
& LEAP & 0.1761 $\pm$ 0.0055 & 0.3105 $\pm$ 0.0012 & 0.1310 $\pm$ 0.0021 & 0.1474 $\pm$ 0.0039 & - & 0.3794 $\pm$ 0.0037 & - \\
& COGNet & 0.2279 $\pm$ 0.0021 & 0.3598 $\pm$ 0.0008 & 0.1391 $\pm$ 0.0031 & 0.1408 $\pm$ 0.0002 & - & 0.4380 $\pm$ 0.0011 & - \\
& RAREMed & 0.2214 $\pm$ 0.0006 & 0.3447 $\pm$ 0.0007 & 0.1440 $\pm$ 0.0053 & 0.1526 $\pm$ 0.0015 & - & 0.4249 $\pm$ 0.0008 & - \\
& ARMR & 0.2280 $\pm$ 0.0032 & 0.3541 $\pm$ 0.0006 & 0.1499 $\pm$ 0.0010 & 0.1337 $\pm$ 0.0037 & - & 0.4347 $\pm$ 0.0022 & - \\
& LAMO & 0.2359 $\pm$ 0.0015 & 0.3582 $\pm$ 0.0040 & 0.1370 $\pm$ 0.0012 & 0.1410 $\pm$ 0.0041 & - & 0.4417 $\pm$ 0.0022 & - \\
\cline{2-9}
& SafeDrug & 0.2107 $\pm$ 0.0058 & 0.3398 $\pm$ 0.0023 & 0.1104 $\pm$ 0.0034 & 0.1473 $\pm$ 0.0037 & - & 0.4183 $\pm$ 0.0041 & - \\
& MoleRec & 0.2311 $\pm$ 0.0029 & 0.3607 $\pm$ 0.0031 & 0.1084 $\pm$ 0.0050 & 0.1391 $\pm$ 0.0022 & - & 0.4424 $\pm$ 0.0033 & - \\
& KELLM & 0.2013 $\pm$ 0.0060 & 0.3164 $\pm$ 0.0006 & 0.1560 $\pm$ 0.0007 & 0.1401 $\pm$ 0.0016 & - & 0.3971 $\pm$ 0.0035 & - \\
& ARC & 0.2297 $\pm$ 0.0019 & 0.3586 $\pm$ 0.0043 & 0.1198 $\pm$ 0.0048 & 0.1374 $\pm$ 0.0037 & - & 0.4398 $\pm$ 0.0026 & - \\
& FLAME & \underline{0.2401 $\pm$ 0.0056} & \underline{0.3644 $\pm$ 0.0049} & 0.1210 $\pm$ 0.0024 & 0.1344 $\pm$ 0.0043 & - & \underline{0.4489 $\pm$ 0.0036} & - \\
\cline{2-9}
& ExpDrug & 0.2257 $\pm$ 0.0022 & 0.3516 $\pm$ 0.0027 & 0.1073 $\pm$ 0.0025 & 0.1410 $\pm$ 0.0038 & - & 0.4342 $\pm$ 0.0024 & - \\
& Reason4Rec & 0.1477 $\pm$ 0.0009 & 0.2480 $\pm$ 0.0029 & 0.1476 $\pm$ 0.0044 & 0.1370 $\pm$ 0.0018 & 0.4207 $\pm$ 0.0030 & 0.3215 $\pm$ 0.0020 & 0.2691 $\pm$ 0.0017 \\
& EXP3RT & 0.2055 $\pm$ 0.0019 & 0.3301 $\pm$ 0.0022 & 0.1331 $\pm$ 0.0032 & 0.1431 $\pm$ 0.0024 & 0.4077 $\pm$ 0.0027 & 0.4086 $\pm$ 0.0021 & 0.3233 $\pm$ 0.0011 \\
\cline{2-9}
& RES-MR-3B & 0.2301 $\pm$ 0.0050 & 0.3340 $\pm$ 0.0009 & \textbf{0.1017 $\pm$ 0.0021} & \underline{0.1303 $\pm$ 0.0007} & \underline{0.4288 $\pm$ 0.0023} & 0.4276 $\pm$ 0.0029 & \underline{0.3403 $\pm$ 0.0021} \\
& RES-MR-8B & \textbf{0.2525 $\pm$ 0.0012}* & \textbf{0.3842 $\pm$ 0.0030}* & \underline{0.1064 $\pm$ 0.0061} & \textbf{0.1247 $\pm$ 0.0034}* & \textbf{0.4439 $\pm$ 0.0045}* & \textbf{0.4682 $\pm$ 0.0019}* & \textbf{0.3708 $\pm$ 0.0015}* \\
\hline

\multirow{15}{*}{eICU}
&LEAP & 0.1831 $\pm$ 0.0031 & 0.3085 $\pm$ 0.0026 & 0.1241 $\pm$ 0.0006 & 0.1364 $\pm$ 0.0005 & - & 0.3833 $\pm$ 0.0021 & - \\
&COGNet & 0.2283 $\pm$ 0.0031 & 0.3580 $\pm$ 0.0012 & 0.1283 $\pm$ 0.0022 & 0.1408 $\pm$ 0.0012 & - & 0.4380 $\pm$ 0.0018 & - \\
& RAREMed & 0.2323 $\pm$ 0.0027 & 0.3601 $\pm$ 0.0027 & 0.1431 $\pm$ 0.0010 & 0.1412 $\pm$ 0.0036 & - & 0.4404 $\pm$ 0.0021 & - \\
& LAMO & 0.2331 $\pm$ 0.0007 & 0.3674 $\pm$ 0.0020 & 0.1173 $\pm$ 0.0012 & 0.1341 $\pm$ 0.0044 & - & 0.4470 $\pm$ 0.0010 & - \\
& ARMR & 0.2249 $\pm$ 0.0008 & 0.3503 $\pm$ 0.0016 & 0.1491 $\pm$ 0.0048 & 0.1401 $\pm$ 0.0024 & - & 0.4305 $\pm$ 0.0012 & - \\
\cline{2-9}
& SafeDrug & 0.2011 $\pm$ 0.0035 & 0.3312 $\pm$ 0.0087 & 0.1104 $\pm$ 0.0034 & 0.1364 $\pm$ 0.0043 & - & 0.4083 $\pm$ 0.0056 & - \\
& MoleRec & 0.2276 $\pm$ 0.0014 & 0.3532 $\pm$ 0.0026 & \underline{0.1065 $\pm$ 0.0039} & 0.1386 $\pm$ 0.0022 & - & 0.4364 $\pm$ 0.0019 & - \\
& KELLM & 0.2197 $\pm$ 0.0048 & 0.3482 $\pm$ 0.0017 & 0.1211 $\pm$ 0.0011 & 0.1374 $\pm$ 0.0022 & - & 0.4282 $\pm$ 0.0034 & - \\
& ARC & 0.2197 $\pm$ 0.0015 & 0.3486 $\pm$ 0.0022 & 0.1201 $\pm$ 0.0008 & 0.1433 $\pm$ 0.0007 & - & 0.4282 $\pm$ 0.0020 & - \\
& FLAME & \underline{0.2397 $\pm$ 0.0043} & 0.3690 $\pm$ 0.0006 & 0.1146 $\pm$ 0.0066 & 0.1311 $\pm$ 0.0029 & - & 0.4519 $\pm$ 0.0021 & - \\
\cline{2-9}
& ExpDrug & 0.2158 $\pm$ 0.0042 & 0.3404 $\pm$ 0.0039 & 0.1086 $\pm$ 0.0038 & 0.1324 $\pm$ 0.0027 & - & 0.4226 $\pm$ 0.0040 & - \\
& Reason4Rec & 0.1877 $\pm$ 0.0015 & 0.2871 $\pm$ 0.0013 & 0.1320 $\pm$ 0.0013 & 0.1423 $\pm$ 0.0049 & \underline{0.4277 $\pm$ 0.0039} & 0.3724 $\pm$ 0.0014 & 0.3053 $\pm$ 0.0011 \\
& EXP3RT & 0.1981 $\pm$ 0.0044 & 0.3017 $\pm$ 0.0043 & 0.1552 $\pm$ 0.0013 & 0.1464 $\pm$ 0.0035 & 0.4119 $\pm$ 0.0049 & 0.3862 $\pm$ 0.0036 & 0.3111 $\pm$ 0.0032 \\
\cline{2-9}
& RES-MR-3B & 0.2376 $\pm$ 0.0009 & \underline{0.3711 $\pm$ 0.0017} & 0.1074 $\pm$ 0.0014 & \underline{0.1291 $\pm$ 0.0046} & 0.4013 $\pm$ 0.0016 & \underline{0.4525 $\pm$ 0.0010} & \underline{0.3462 $\pm$ 0.0007} \\
& RES-MR-8B & \textbf{0.2593 $\pm$ 0.0022}* & \textbf{0.3908 $\pm$ 0.0039}* & \textbf{0.1009 $\pm$ 0.0009}* & \textbf{0.1247 $\pm$ 0.0016}* & \textbf{0.4368 $\pm$ 0.0016}* & \textbf{0.4758 $\pm$ 0.0021}* & \textbf{0.3727 $\pm$ 0.0016}* \\
\hline
\end{tabular}
\vspace{-2mm}
\end{table*}

\noindent\textbf{Metrics.} We evaluate our method from three perspectives: recommendation accuracy, medication safety, and reasoning interpretability. We adopt the commonly-used accuracy metrics, i.e., \textbf{Jaccard Similarity (Jaccard)}~\cite{survey} and \textbf{F1-score (F1)}~\cite{survey2}, to measure the overlap between the predicted set $\hat{\mathbf{m}}_i$ and the ground-truth set $\mathbf{m}_i$ at the last visit for each patient $i$ in the test set $\mathcal{U}$:
\begin{equation}
    \text{Jaccard} = \frac{1}{|\mathcal{U}|} \sum_{i=1}^{|\mathcal{U}|} \frac{|\mathbf{m}_i \cap \hat{\mathbf{m}}_i|}{|\mathbf{m}_i \cup \hat{\mathbf{m}}_i|},\ 
    \text{F1} = \frac{1}{|\mathcal{U}|} \sum_{i=1}^{|\mathcal{U}|} \frac{2 \cdot \text{Pre}_i \cdot \text{Rec}_i}{\text{Pre}_i + \text{Rec}_i},
\end{equation}
where $\text{Pre}_i = \frac{|\hat{\mathbf{m}}_i \cap \mathbf{m}_i|}{|\hat{\mathbf{m}}_i|}$ and $\text{Rec}_i = \frac{|\hat{\mathbf{m}}_i \cap \mathbf{m}_i|}{|\mathbf{m}_i|}$.
Safety is assessed via the \textbf{DDI rate}~\cite{GAMENET} and our proposed \textbf{Risk Intrusion Score (RIS)}. The DDI rate calculates the proportion of medication pairs in $\hat{\textbf{m}}_i$ that induce interactions based on the adjacency matrix $\mathbf{M}_{ddi}$:
\begin{equation}
\text{DDI} = \frac{1}{|\mathcal{U}|} \sum_{i=1}^{|\mathcal{U}|}
\frac{
\sum_{j=1}^{|\hat{\mathbf{m}}_i|}\sum_{k=j+1}^{|\hat{\mathbf{m}}_i|}\mathbb{I}\!\left(\mathbf{M}_{ddi}\!\left[\hat{\mathbf{m}}_i^{(j)},\,\hat{\mathbf{m}}_i^{(k)}\right]=1\right)
}{
\sum_{j=1}^{|\hat{\mathbf{m}}_i|}\sum_{k=j+1}^{|\hat{\mathbf{m}}_i|} 1
},
\end{equation}
where $\mathbb{I}(\cdot)$ is the indicator function. To empirically assess patient-specific risk tolerance, which is not directly observable from EHRs, we define \textbf{RIS} as a proxy metric~\cite{proxy1} that measures the extent to which predicted drugs violate each patient's risk boundary $B_i$:
\begin{equation}
    \text{RIS} = \frac{1}{|\mathcal{U}|} \sum_{i=1}^{|\mathcal{U}|} \frac{1}{|\hat{\mathbf{m}}_i|} \sum_{d \in \hat{\mathbf{m}}_i} \frac{
    \max\left(0,  s(\boldsymbol{\mu}_{\text{risk}}(d), \mathbf{v}_{\text{neg}}) - B_i \right)}
    {1 - B_i}.
\end{equation}
Grounded in external medical knowledge, RIS leverages frozen disentangled factors as a safety signal complementary to DDI, capturing patient-specific risks. 
To evaluate the quality of explanations provided by reasoning trajectory, we follow previous work~\cite{Reason4Rec, grefer, xrec} and adopt BLEURT. 
Specifically, \textbf{BLEURT} measures the semantic alignment between the generated reasoning paths and reference explanations~\cite{BLEURT}. 
Meanwhile, we introduce two composite metrics based on harmonic means to measure the pair-wise trade-offs between accuracy, safety, and explainability: \textbf{$F_{AS}$} (Accuracy-Safety) and \textbf{$F_{AE}$} (Accuracy-Explainability):
\begin{equation}
    F_{AS} = \frac{2 \cdot S_{Acc} \cdot S_{Safe}}{S_{Acc} + S_{Safe}}, \quad F_{AE} = \frac{2 \cdot S_{Acc} \cdot \text{BLEURT}}{S_{Acc} + \text{BLEURT}},
\end{equation}
where $S_{Acc} = \frac{\text{Jaccard} + \text{F1}}{2}$ and $S_{Safe} = \frac{(1-\text{DDI}) + (1-\text{RIS})}{2}$.

\noindent \textbf{Baseline Methods.} To comprehensively evaluate RES-MR, we select representative and state-of-the-art baselines and group them into three categories: 
(1) \textbf{General medication recommendation methods:} LEAP~\cite{LEAP}, COGNet~\cite{COG}, RAREMed~\cite{RAREMed}, ARMR~\cite{armr}, LAMO~\cite{LAMO}; 
(2) \textbf{Safety methods:} SafeDrug~\cite{safedrug}, MoleRec~\cite{molerec}, ARC~\cite{ARC}, KELLM~\cite{KELLM}, FLAME~\cite{FLAME}; and 
(3) \textbf{Explainable methods:} ExpDrug~\cite{ExpDrug}, Reason4Rec~\cite{Reason4Rec}, EXP3RT~\cite{EXP3RT}. 
Among them, the \textbf{LLM-based methods} include KELLM, LAMO, FLAME, Reason4Rec, and EXP3RT, while the remaining baselines are conventional methods.
To ensure a rigorous and fair comparison, we employ MICRON~\cite{micron} as the backbone of ARC, following the best practice reported in the original paper~\cite{ARC}. 
For LLM-based methods, we use Llama-3.1-Aloe-Beta-8B~\cite{aloe} as backbone to control for foundation model capacity. 
Moreover, we transfer Reason4Rec and EXP3RT to our medication recommendation setting.

\noindent \textbf{Implementation Details.} 
All experiments are conducted on a cluster equipped with 8 NVIDIA A100-SXM4 GPUs, Intel Xeon Platinum 8370C CPUs, and 512GB RAM. 
For the \textbf{Dual-Factor Disentanglement module}, the drug encoder is configured with latent/shared dimensions of 128/512 and pretrained for 100 epochs using Adam with a learning rate of $10^{-4}$. After pretraining, we freeze the encoder as a fixed feature extractor.
We search $\lambda_{\text{orth}}$ in $[0.05,0.50]$ with step size $0.05$ and $\beta_{\text{KL}}$ in $\{1,3,5,7,9\}\times 10^{-4}$. 
For \textbf{Knowledge Graph Retrieval}, we leverage PrimeKG~\cite{primekg}, performing 1-hop neighbor expansion via Neo4j and employing bge-large-en-v1.5~\cite{bgeembedding} to retrieve the top-5 entities relevant to clinical notes.
In \textbf{Stage 1}, we employ GPT-5 as the teacher model and Llama-3.1-Aloe-Beta-8B~\cite{aloe} as the student model. 
We set the context length to 8,192 and retain only the latest three visits to limit input length while preserving space for retrieved knowledge.
We fine-tune the student via LoRA (Rank $r=8$, Alpha $\alpha=16$, Dropout 0.05). The training uses a learning rate of $2\times10^{-4}$, the AdamW optimizer, and Flash Attention 2~\cite{flashattention2}. 
In \textbf{Stage 2}, we initialize with Stage 1 weights, maintaining LoRA settings, and employ GSPO~\cite{GSPO} ($G=4, \epsilon=0.2$) with a learning rate of $10^{-6}$. The risk boundary strictness $\delta$ is set to 0.9, and reward weights $\lambda_{\{j, e, d, r\}}$ are tuned within $\{1, 3, 5, 7, 10\}$. The adaptive gating threshold $\tau_t$  is linearly annealed from 0.2 during training. 
All experiments are repeated five times with different random seeds.

\subsection{Overall Performance Comparison}

We evaluate RES-MR on the MIMIC-III dataset across accuracy, safety, and explainability. Meanwhile, following~\cite{FLAME, sttrl}, we evaluate the generalization and transferability of our RES-MR by initializing the model with SFT weights from MIMIC-III and performing only one epoch of RAPO training on MIMIC-IV and eICU datasets, whereas baselines are fully trained on these datasets. We implement RES-MR using both Llama-3.2-3B-Instruct and Llama-3.1-Aloe-Beta-8B backbones to assess performance across different parameter scales, denoted as RES-MR-3B and RES-MR-8B. 

\footnotetext[1]{BLEURT and $F_{AE}$ are only computed for methods that generate textual explanations; ``--'' indicates that the method does not output explanation text.}

As shown in Table~\ref{tab:overall}, we report the average performance and standard deviations of each method on three datasets. From these results, we observe the following: 
1) RES-MR consistently achieves the best overall performance across three datasets. 
In particular, RES-MR-8B outperforms all baselines, achieving a 7.7\% improvement in Jaccard over the state-of-the-art FLAME on MIMIC-III. These gains are attributed to our ECRD, which fosters explicit clinical reasoning rather than mere co-occurrence patterns, and to RAPO, which dynamically optimizes the efficacy-safety trade-off. Notably, under our transfer setting where RES-MR is initialized from MIMIC-III SFT weights and only lightly adapted on MIMIC-IV and eICU, it still maintains superior performance, demonstrating robust transferability.
2) Safety baselines can be overly conservative, making it difficult to balance accuracy and safety and leading to suboptimal performance. For instance, SafeDrug and MoleRec primarily enforce safety via DDI penalties. Consequently, these methods may prioritize interaction avoidance over therapeutic coverage, resulting in overly restricted medication combinations. In contrast, RES-MR achieves the lowest RIS and DDI rates while maintaining superior accuracy. Since RIS is measured with respect to personalized risk boundaries, these results suggest the effectiveness of RAPO in better balancing therapeutic efficacy and safety under patient-specific safety constraints. 
3) General reasoning baselines (e.g., Reason4Rec and EXP3RT) provide recommendation rationales but show limited overall performance. While these methods introduce interpretability, they fail to translate their reasoning capabilities into clinically accurate and safe decisions. This limitation may stem from the fact that these frameworks are adapted from general domains that prioritize user preference, and consequently lack explicit modeling of stringent safety constraints and heterogeneous clinical contexts.

\begin{table}[ht]
\renewcommand{\arraystretch}{0.9}
\small
\centering
\caption{Ablation study of RES-MR on the MIMIC-III dataset.}
\vspace{-2mm}
\label{tab:ablation}
\begin{tabular}{l|ccc}
\hline
Methods 
& Jaccard $\uparrow$
& DDI $\downarrow$
& BLEURT $\uparrow$ \\
\hline
RES-MR w/o Reasoning & 0.2414 & 0.1474 & - \\
RES-MR w/o Retrieval & 0.2244 & 0.1626 & 0.4117 \\
RES-MR w/o RAPO & 0.2668 & 0.1505 & 0.4408 \\
RES-MR w/o Disentangle & 0.2631 & 0.1211 & 0.4410 \\
RES-MR w/o Dense Reward & 0.2714 & 0.1103 & 0.4483 \\
RES-MR w/o Sparse Reward & 0.2631 & 0.1386 & 0.4396 \\
RES-MR w/o Trade-off & 0.2703 & 0.1324 & 0.4387 \\
RES-MR w/o Clipping & 0.2688 & 0.1086 & 0.4337 \\
\textbf{RES-MR-8B} & \textbf{0.2803} & \textbf{0.1031} & \textbf{0.4538} \\
\hline
\end{tabular}
\vspace{-2mm}
\end{table}

\subsection{Ablation Study}
To validate the effectiveness of each component in RES-MR, we design the following variants for comparative analysis: 
1) w/o Reasoning: Fine-tunes on labels only, skipping the reasoning stage;
2) w/o Retrieval: Removes retrieved context from the input;
3) w/o RAPO: Removes the reinforcement learning stage;
4) w/o Disentangle: Uses raw LLM embeddings without dual-factor disentanglement;
5) w/o Dense/Sparse Reward: Removes dense (Efficacy/Risk) or sparse (Jaccard/DDI) rewards;
6) w/o Trade-off: Uses a fixed static sum instead of a dynamic trade-off;
7) w/o Clipping: Removes the reward clipping. 
The experimental results in Table~\ref{tab:ablation} indicate several key findings.
First, significant accuracy drops for w/o Retrieval and w/o Reasoning confirm that retrieval-augmented reasoning is fundamental for understanding complex clinical contexts and making accurate decisions. 
Second, the marked increase in DDI rates for w/o RAPO demonstrates that supervised learning alone is insufficient to internalize patient-specific risk boundary, necessitating RL for exploration and correction. 
Furthermore, the comprehensive performance decline in w/o Disentangle highlights that disentangled factors are critical for constructing high-quality prototypes. 
Finally, removing specific reward components or the Trade-off/Clipping mechanisms hinders the model from achieving an optimal balance between efficacy and safety (e.g., w/o Clipping may induce reward hacking, ignoring safety risks for high scores), highlighting the importance of our fine-grained reward design.

\begin{figure}[t]
    \centering
    \captionsetup[subfigure]{font=footnotesize, labelfont=normalfont}
    \begin{subfigure}[t]{0.23\textwidth}
        \centering
        \includegraphics[width=\textwidth]{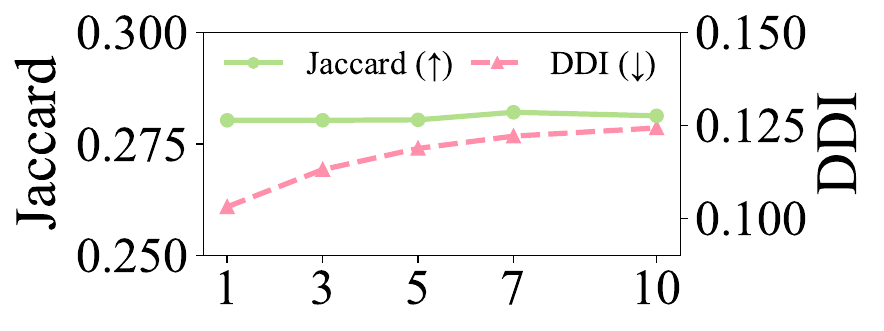}
        \vspace{-6mm}
        \caption{$\lambda_j$}
        \label{fig:reward_j}
    \end{subfigure}
    \begin{subfigure}[t]{0.23\textwidth}
        \includegraphics[width=\textwidth]{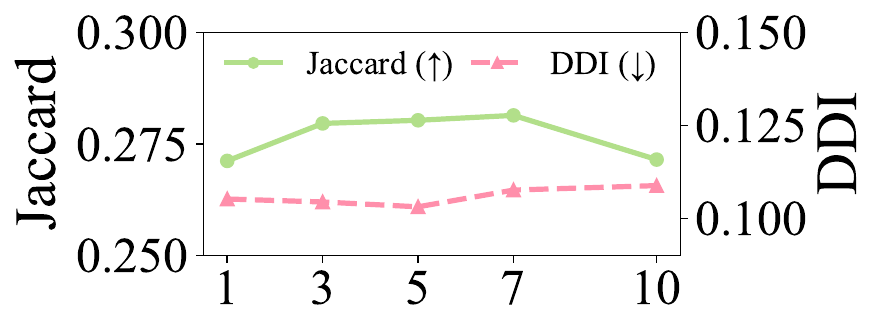}
        \vspace{-6mm}
        \caption{$\lambda_e$}
        \label{fig:reward_e}
    \end{subfigure}
    \begin{subfigure}[t]{0.23\textwidth}
        \centering
        \includegraphics[width=\textwidth]{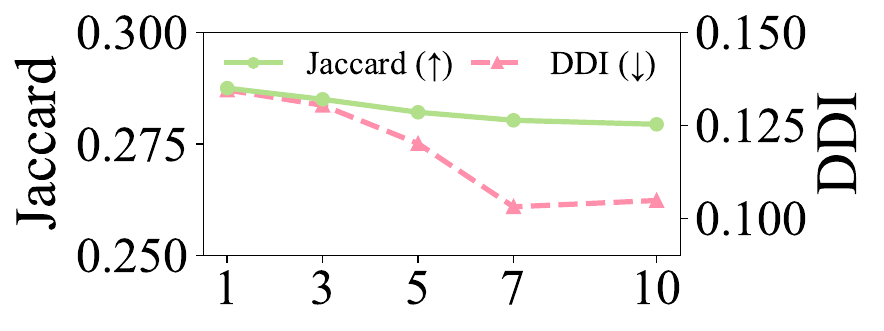}
        \vspace{-6mm}
        \caption{$\lambda_d$}
        \label{fig:reward_d}
    \end{subfigure}
    \begin{subfigure}[t]{0.23\textwidth}
        \includegraphics[width=\textwidth]{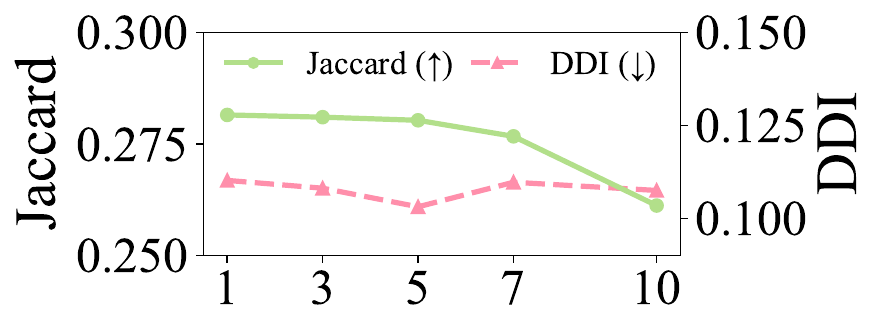}
        \vspace{-6mm}
        \caption{$\lambda_r$}
        \label{fig:reward_r}
    \end{subfigure}
\caption{Hyperparameter effects on model performance.}
\label{fig:hyperparameters}
\end{figure}

\subsection{Hyperparameter Analysis}
In this section, we analyze the impact of four key hyperparameters on the performance of RES-MR on the MIMIC-III dataset, i.e., Jaccard reward weight $\lambda_j$, efficacy bonus weight $\lambda_e$, DDI penalty weight $\lambda_d$, and risk penalty weight $\lambda_r$. We vary these parameters within the range of $\{1, 3, 5, 7, 10\}$ and report the results of Jaccard and DDI rate in Figure~\ref{fig:hyperparameters}. 
As shown in Figure~\ref{fig:hyperparameters} (a) and (b), increasing efficacy-related weights generally encourages the model to better fit the target medication distribution. However, we observe that an excessively large $\lambda_e$ leads to a slight decline in Jaccard. This suggests that over-emphasizing semantic similarity may divert the model towards ``functionally similar'' alternatives rather than the exact ground-truth labels, causing a drop in performance. 
Figure~\ref{fig:hyperparameters} (c) and (d) illustrate the trade-off between accuracy and safety. As $\lambda_d$ and $\lambda_r$ increase, DDI rates decrease significantly, validating the effectiveness of RAPO in suppressing unsafe behaviors. Notably, extremely high penalties cause a sharp collapse in Jaccard, reflecting an over-conservative policy that refuses prescriptions to avoid potential risks. Given that clinical safety is our primary consideration, we select weight settings that strictly control risks, even though this may incur a marginal loss in accuracy.

\begin{figure}[t]
    \centering
    \includegraphics[width=0.9\linewidth]{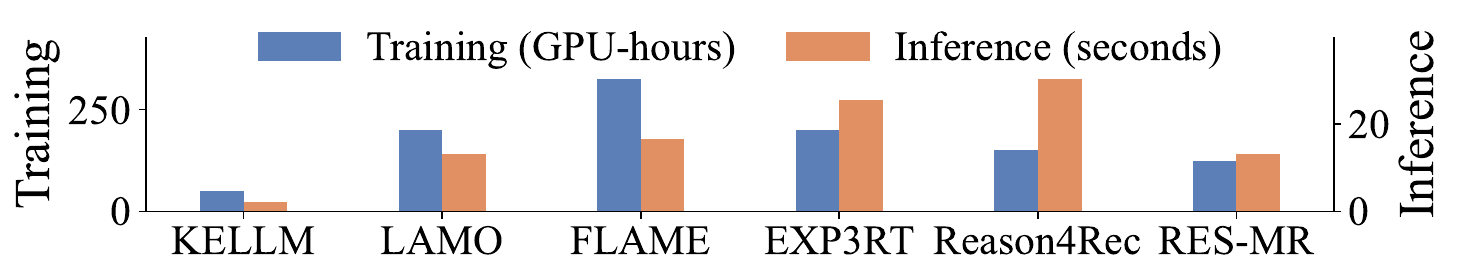}
    \vspace{-2mm}
    \caption{Efficiency analysis of training cost and inference time on MIMIC-III. Training uses $8\times$ NVIDIA A100 GPUs, while inference uses a single NVIDIA A100 GPU.}
    \label{fig:efficiency}
    \vspace{-2mm}
\end{figure}

\subsection{Efficiency Analysis}
To assess practical deployability, we compare the computational costs of RES-MR with representative LLM-based baselines in Figure~\ref{fig:efficiency}, with all experiments conducted in a unified hardware environment. While KELLM achieves the lowest time due to its non-generative discriminative design, RES-MR demonstrates superior efficiency among generative LLM-based methods. Specifically, regarding training, RES-MR requires approximately 125 GPU-hours, significantly lower than FLAME, which relies on a resource-intensive multi-stage training pipeline. In terms of inference, unlike reasoning-enhanced baselines such as Reason4Rec and EXP3RT that incur high time overheads due to generating lengthy rationales, RES-MR achieves an inference time of only 13.3~s, comparable to the non-reasoning baseline LAMO. This efficiency is primarily attributed to the ECRD stage, which distills concise, clinically grounded reasoning paths, avoiding costly multi-pass generation during inference.

\begin{figure}[t]
    \centering
    \includegraphics[width=0.9\linewidth]{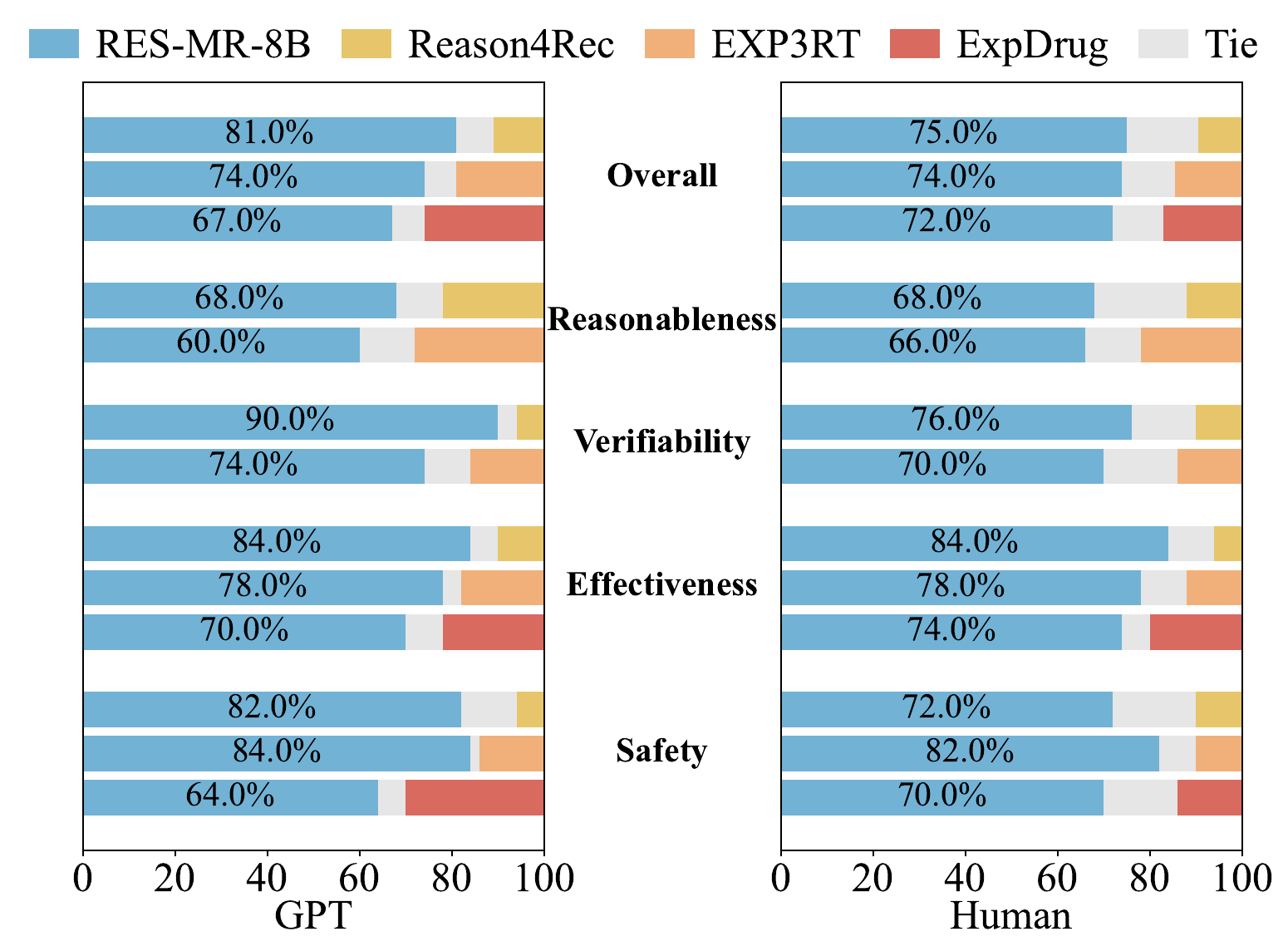}
    \vspace{-2mm}
    \caption{Human and GPT evaluations on pairwise quality comparison of explanations (p-value < 0.05).}
    \label{fig:explanation}
    \vspace{-8pt}
\end{figure}

\subsection{Explanation Quality Assessment}
To evaluate explanation quality, we conduct a blind pairwise comparison on 50 randomly sampled MIMIC-III test instances, assessed by both GPT-5 and three human evaluators with medical or pharmaceutical backgrounds. 
The evaluation assesses the generated explanations via \textbf{Reasonableness} and \textbf{Verifiability}, and the recommended medication sets via \textbf{Effectiveness} and \textbf{Safety}. Note that ExpDrug is evaluated only on the latter two due to its lack of textual explanations.
As shown in Figure~\ref{fig:explanation}, RES-MR consistently outperforms baselines. This superiority stems from ECRD, which enables the model to internalize verifiable, clinically grounded reasoning paths, and RAPO, which explicitly enforces safety via reward shaping. Moreover, human evaluators show substantial inter-rater agreement across all four dimensions (Fleiss' $\kappa > 0.7$), and the agreement between human and GPT-5 judgments further supports the reliability of these findings.

\begin{table}[t]
\renewcommand{\arraystretch}{0.9}
\small
\centering
\caption{A case study of a patient in the MIMIC-III dataset.}
\label{tab:case_study}
\resizebox{\columnwidth}{!}{
\begin{tabular}{p{0.25\columnwidth} p{0.70\columnwidth}}
\toprule
\multicolumn{2}{l}{\textbf{Subject 41115} (80y, Male). \textbf{Proc}: AVR + CABG. \textbf{Hx}: AFib, HTN.} \\
\midrule
\textbf{Context} & \textbf{Verifiable Rationale $\rightarrow$ Recommendation} \\
\midrule
\textbf{Surgical \newline Protocol} & 
\textit{Expl:} ``Given \textbf{extracorporeal circulation}, requires \textbf{reversal of neuromuscular blockade}...'' \newline
$\hookrightarrow$ \textbf{Rec}: \textbf{Neostigmine}, \textbf{Glycopyrronium} \\
\cmidrule(l){2-2}

\textbf{Stroke \newline Risk} & 
\textit{Expl:} ``Restart Warfarin post-op \textbf{once hemostasis secured} to prevent thromboembolism.'' \newline
$\hookrightarrow$ \textbf{Rec}: \textbf{Warfarin} (\textcolor{red}{\textit{personalized risk boundary}}) \\
\cmidrule(l){2-2}

\textbf{Fluid \newline Balance} & 
\textit{Expl:} ``Replete K+ losses caused by \textbf{aggressive diuresis} to maintain electrical stability.'' \newline
$\hookrightarrow$ \textbf{Rec}: \textbf{Furosemide}, \textbf{Potassium Cl} \\
\bottomrule
\end{tabular}
}
\end{table}

\subsection{Case Study}
Table~\ref{tab:case_study} exemplifies how RES-MR generates medication recommendations with \textbf{verifiable explanations}. Instead of relying on superficial co-occurrence, the model demonstrates deep reasoning: it adheres to procedural protocols by recommending Neostigmine and Glycopyrronium for neuromuscular blockade reversal, while leveraging pharmacological knowledge from \textit{Knowledge Graph Retrieval} to anticipate Furosemide-induced hypokalemia, proactively co-prescribing Potassium Chloride. Furthermore, in addressing the conflict between stroke risk (AFib) and postoperative bleeding, RES-MR leverages fine-grained risk perception from RAPO to establish a \textbf{patient-specific risk boundary}. It prudently advises prescribing Warfarin only ``once hemostasis is secured.'' This nuance attests to the model's ability to handle intricate clinical trade-offs while generating verifiable rationales and accurate, safe recommendations. 

\section{Conclusion}
In this paper, we proposed RES-MR, a novel risk-aware reasoning framework designed to address the critical challenges of interpretability and personalized safety in medication recommendation. 
By integrating Explainable Clinical Reasoning Distillation (ECRD) for transparent reasoning and Risk-Aware Policy Optimization (RAPO) for dynamic safety balancing, RES-MR transforms recommendations into an explainable generative process. 
Experiments on three datasets demonstrate that RES-MR significantly outperforms state-of-the-art baselines, achieving a superior accuracy-safety trade-off. 
In future work, we will explore agentic reasoning workflows with iterative retrieval and self-verification to strengthen faithful clinical reasoning and further enhance decision reliability.

\begin{acks}
This work is supported by the National Natural Science Foundation of China (No. 62376130), Program of New Twenty Policies for Universities of Jinan (No.202333008), and the Pilot Project for Integrated Innovation of Science, Education, Industry of Qilu University of Technology (Shandong Academy of Sciences) (No.2025ZDZX01).
\end{acks}

\bibliographystyle{ACM-Reference-Format}
\balance
\bibliography{sample-base}

\end{document}